%% file: main.tex
\documentclass[10pt,journal,compsoc]{IEEEtran}
\usepackage{tikz}
\usepackage{times}
\usepackage{epsfig}
\usepackage{graphicx}
\usepackage{amsmath,amssymb,bm}
\usepackage{color,colortbl}
\usepackage{diagbox}
\usepackage{caption}
\usepackage{comment}
\usepackage{url}            
\usepackage{amsfonts}       
\usepackage{nicefrac}       
\usepackage{microtype}      
\usepackage{multirow}
\usepackage{cuted}
\usepackage{booktabs}
\usepackage{multirow,textcomp}
\usepackage[ruled,vlined]{algorithm2e}
\usepackage[breaklinks=true,bookmarks=false]{hyperref}

\ifCLASSOPTIONcompsoc
  \usepackage[nocompress]{cite}
\else
  \usepackage{cite}
\fi

\usepackage[capitalize]{cleveref}
\crefname{section}{Sec.}{Secs.}
\Crefname{section}{Section}{Sections}
\Crefname{table}{Table}{Tables}
\crefname{table}{Tab.}{Tabs.}

\input{macros}

\begin{document}

\title{Extracting Semantic Knowledge from GANs with Unsupervised Learning}

\author{Jianjin~Xu,
        Zhaoxiang~Zhang,
        Xiaolin~Hu,~\IEEEmembership{Senior Member,~IEEE,}
\IEEEcompsocitemizethanks{
\IEEEcompsocthanksitem J. Xu is with the School of Mathematics and Computer Science, Panzhihua University and the Department of Computer Science and Technology, Tsinghua University.\protect\\
E-mail: xujj15@gmail.com
\IEEEcompsocthanksitem Z. Zhang is with with the Insitute of Automation, Chinese Academy of Sciences and the Centre for Artificial Intelligence and Robotics, Hong Kong Innovation and Science Insitute, Chinese Academy of Sciences.\protect\\
E-mail: zhaoxiang.zhang@ia.ac.cn
\IEEEcompsocthanksitem X. Hu is with the Department of Computer Science and Technology, Institute for Artificial Intelligence, State Key Laboratory of Intelligent Technology and Systems, THU-Bosch JCML Center, THBI, IDG/McGovern Institute for Brain Research, BNRist, Tsinghua University, Beijing, China. He is also with the Chinese Institute for Brain Research (CIBR), Beijing, China.\protect\\
E-mail: xlhu@tsinghua.edu.cn}
\thanks{This work was supported in part by the National Natural Science Foundation of China (Nos. 62061136001, 61836014, and U19B2034) and THU-Bosch JCML center. Corresponding author: Xiaolin Hu.}}

\IEEEtitleabstractindextext{%
\begin{abstract}
  Recently, unsupervised learning has made impressive progress on various tasks.
  Despite the dominance of discriminative models, increasing attention is drawn to representations learned by generative models and in particular, Generative Adversarial Networks (GANs).
  Previous works on the interpretation of GANs reveal that GANs encode semantics in feature maps in a linearly separable form.
  In this work, we further find that GAN's features can be well clustered with the linear separability assumption.
  We propose a novel clustering algorithm, named KLiSH, which leverages the linear separability to cluster GAN's features.
  KLiSH succeeds in extracting fine-grained semantics of GANs trained on datasets of various objects, \eg, car, portrait, animals, and so on.
  With KLiSH, we can sample images from GANs along with their segmentation masks and synthesize paired image-segmentation datasets.
  Using the synthesized datasets, we enable two downstream applications.
  First, we train semantic segmentation networks on these datasets and test them on real images, realizing unsupervised semantic segmentation.
  Second, we train image-to-image translation networks on the synthesized datasets, enabling semantic-conditional image synthesis without human annotations.
\end{abstract}

}

\maketitle

\input{intro}
\input{related}
\input{cluster}
\input{expr}
\input{application}
\input{conclusion}

\bibliographystyle{IEEEtran}
\bibliography{bib}

\begin{IEEEbiography}[{\includegraphics[width=1in,height=1.25in,clip,keepaspectratio]{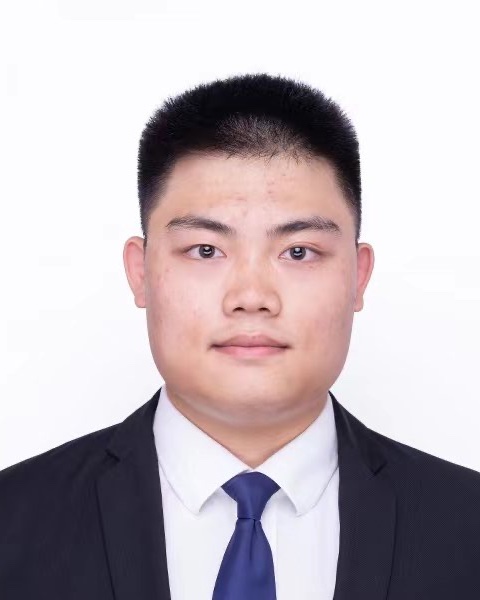}}]{\textbf{Jianjin Xu}} (S'01-M'08-SM'13) received a B.E. degree in computer science from the Tsinghua University, Beijing, China, in 2019, and a M.S. degree in computer science from the Columbia University, New York City, NY, U.S., in 2021. He is currently an Assistant Research Scientist at the School of Mathematics and Computer Science, Panzhihua University, Sichuan, China. His current research interests include generative models, neural network interpretation, and computer vision. Previously he was a research assistant at the Department of Computer Science and Technology, Tsinghua University.
\end{IEEEbiography}

\begin{IEEEbiography}[{\includegraphics[width=1in,height=1.25in,clip,keepaspectratio]{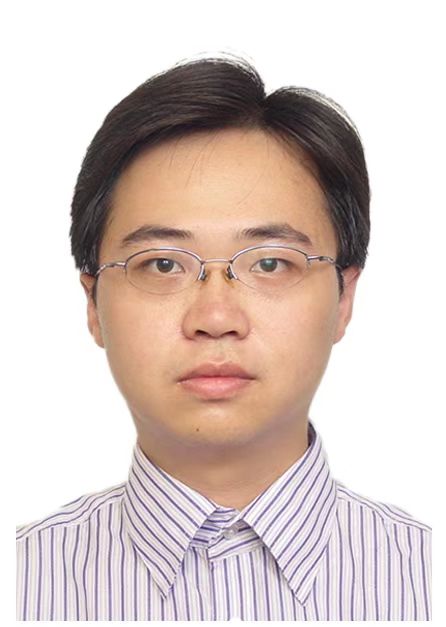}}]{\textbf{Zhaoxiang Zhang}} received his bachelor's degree in Circuits and Systems from the University of Science and Technology of China (USTC) in 2004. In 2004, he joined the National Laboratory of Pattern Recognition (NLPR), Institute of Automation, Chinese Academy of Sciences, under the supervision of Professor Tieniu Tan, and he received his Ph.D. degree in 2009. In October 2009, he joined the School of Computer Science and Engineering, Beihang University, as an Assistant Professor (2009-2011), an Associate professor (2012-2015) and the vise-director of the Department of Computer application technology (2014-2015). In July 2015, he returned to the Institute of Automation, Chinese Academy of Sciences. He is now a full Professor in the Center for Research on Intelligent Perception and Computing (CRIPAC) and the National Laboratory of Pattern Recognition (NLPR). His research interests include Computer Vision, Pattern Recognition, and Machine Learning. Recently, he specifically focuses on biologically inspired intelligent computing  and its applications on human analysis and scene understanding. He has published more than 150 papers in the international journals and conferences, including reputable international journals such as IEEE TPAMI, IJCV, JMLR and top level international conferences like CVPR, ICCV, ECCV, ICLR, NeurIPS, AAAI and IJCAI. He served as the associated editor of IEEE TCSVT, PR, and Frontiers of Computer Science. He served as the area chair of international conferences like CVPR, ICCV, AAAI, IJCAI. He is a senior member of IEEE.
\end{IEEEbiography}

\begin{IEEEbiography}[{\includegraphics[width=1in,height=1.25in,clip,keepaspectratio]{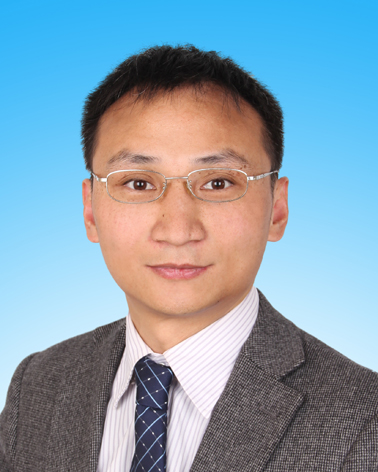}}]{\textbf{Xiaolin Hu}} (S'01-M'08-SM'13) received B.E. and M.E. degrees in automotive engineering from the Wuhan University of Technology, Wuhan, China, in 2001 and 2004, respectively, and a Ph.D. degree in automation and computer-aided engineering from the Chinese University of Hong Kong, Hong Kong, in 2007. He is currently an Associate Professor at the Department of Computer Science and Technology, Tsinghua University, Beijing, China. His current research interests include deep learning and computational neuroscience. At present, he is an Associate Editor of the IEEE Transactions on Pattern Analysis and Machine Intelligence, IEEE Transactions on Image Processing, and Cognitive Neurodynamics. Previously he was an Associate Editor of the IEEE Transactions on Neural Networks and Learning Systems.
\end{IEEEbiography}

\end{document}

%% file: macros.tex
\newtheorem{definition}{Definition}

\makeatletter
\DeclareRobustCommand\onedot{\futurelet\@let@token\@onedot}
\def\onedot{\ifx\@let@token.\else.\null\fi\xspace}

\def\eg{\emph{e.g}\onedot} 
\def\ie{\emph{i.e}\onedot} 
 
\def\etc{\emph{etc}\onedot} 
 
\def\etal{\emph{et al}\onedot}
\makeatother

\newcommand{\norm}[1]{\left\lVert#1\right\rVert}

\newcommand{\bR}[0]{\mathbb{R}}

\newcommand{\W}[0]{\mathbf{W}}

\newcommand{\x}[0]{\mathbf{x}}

\newcommand{\mA}[0]{\mathcal{A}}

\newlength\savedwidth
\newcommand\whline[1]{\noalign{\global\savedwidth\arrayrulewidth
                               \global\arrayrulewidth #1} %
                      \hline
                      \noalign{\global\arrayrulewidth\savedwidth}}

\makeatletter
\renewcommand{\paragraph}{%
  \@startsection{paragraph}{5}%
  {\z@}{0.60ex \@plus 1ex \@minus .15ex}{-0.8em}%
  {\normalfont\normalsize\bfseries}%
}
\makeatother

%% file: intro.tex
\section{Introduction}

\IEEEPARstart{R}{e}presentation is at the core of modern deep learning practice.
Bengio \etal \cite{bengio2013representation} propose that a good representation should have (1) \emph{explanatory factors organized in hierarchy}, and (2) \emph{natural clustering of categories}.
Chan \etal \cite{chan2022redunet} claim that the objective of a deep network is to learn a linearly discriminative representation of the data.
Researchers have found that the representation learned by GANs \cite{goodfellow2014generative} not only explains the variation of images \cite{shen2020interfacegan,yang2021semantic,bau2018gan} but also is linearly discriminative \cite{xu2021linear}, indicating that GANs learn a good representation.

Given a dataset of images, a GAN learns to model the distribution of data and synthesizes images unseen in the dataset.
The image synthesis process of GAN is typically a mapping from noise vectors to a series of intermediate feature maps and then to RGB images.
From the perspective of causality, GAN's representations contain all the information in the image.
In other words, the variation factors (\eg, the attributes of objects) present in the image must also exist in GAN's representation.
Previous works have identified two types of variations captured by GANs.
First, the image-level attributes (\eg, the gender of face images) are separable by linear hyperplanes in GAN's latent space \cite{shen2020interfacegan,jahanian2019steerability,yang2021semantic,voynov2020unsupervised,shen2021closed}.
Second, the object-level semantics (\eg, the presence and location of a window in bedroom images) are encoded in GAN's feature maps \cite{bau2018gan} in a linearly separable form \cite{xu2021linear}.
We term the rich semantic information in GAN's features as the semantic knowledge learned by the GANs.

\begin{figure}[t]
    \centering
    \includegraphics[width=0.98\linewidth]{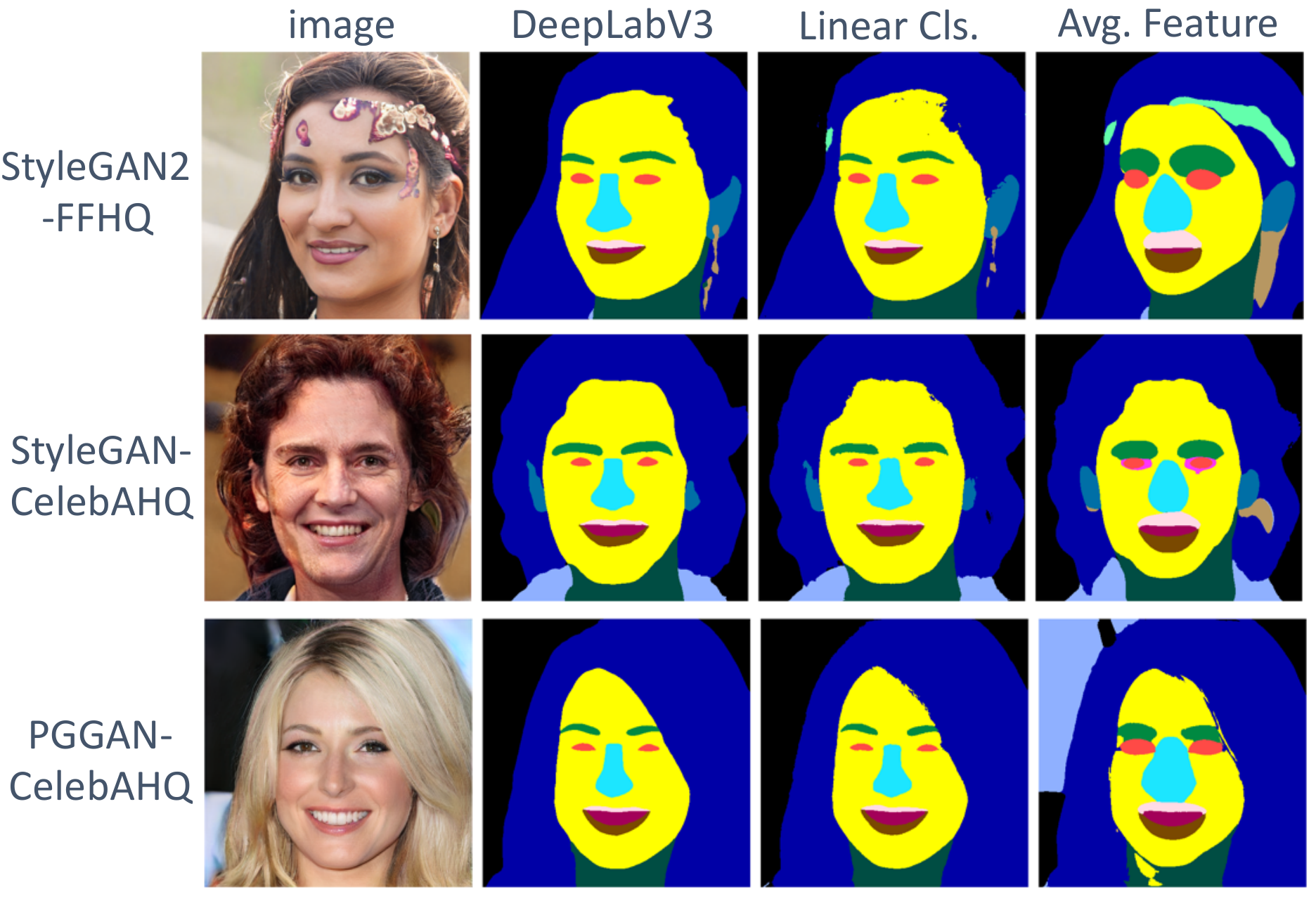}
    \caption{Semantic segmentation obtained by using the average feature of each class compared to using a trained linear classifier. The prediction from a pretrained DeepLabV3 \cite{chen2017rethinking} is regarded as the groundtruth. The linear classifier is supervised by a DeepLabV3 network pretrained on CelebAMask-HQ dataset \cite{lee2020maskgan} according to Xu \etal \cite{xu2021linear}.}
    \label{fig:seg_centroid}
\end{figure}

Existing works \cite{ling2021editgan,zhang2021datasetgan,tritrong2021repurposing} often train few-shot learning classifiers to reveal GAN's semantic knowledge.
The classifiers predict each spatial location of GAN's feature maps.
Training the classifiers needs only a few labeled examples.
The trained classifier is subsequently used for image editing \cite{ling2021editgan,zhang2021datasetgan} and semantic segmentation \cite{tritrong2021repurposing}.
However, the labeling cost for the few-shot learning method is still high.
As reported in \cite{ling2021editgan,zhang2021datasetgan}, the number of annotations ranges from 16 to 40 and labeling each image takes around 10 minutes even for a skillful user.
Therefore, if one can design an unsupervised segmentation algorithm based on GAN's feature maps for image editing, it would enable users to upload their own images, train their own GANs, and edit those images without much labor cost.
This is the aim of this study.
Specifically, we study how to reveal GAN's semantic knowledge with unsupervised learning methods.

Existing unsupervised learning methods mostly apply conventional clustering algorithms on GAN's features, \eg, K-means \cite{Collins20,kafri2022stylefusion} and Agglomerative Hierarchical Clustering (AHC) \cite{zheng2021unsupervised}.
They use clustering results as an intermediate step for editing the attribute \cite{kafri2022stylefusion} or swapping the style \cite{Collins20} in a localized region.
However, the obtained clusters are not satisfactory.
Please refer to their results \cite{Collins20,kafri2022stylefusion,zheng2021unsupervised} and also our reproduction in \cref{fig:klish_results}.
For example, for face images, we observed the oversegmentation of the hair class, and the missing of tiny classes like brow, upper lip, and lower lip (see \cref{fig:klish_results}, the last column).

\begin{figure}[t]
    \centering
    \includegraphics[width=0.98\linewidth]{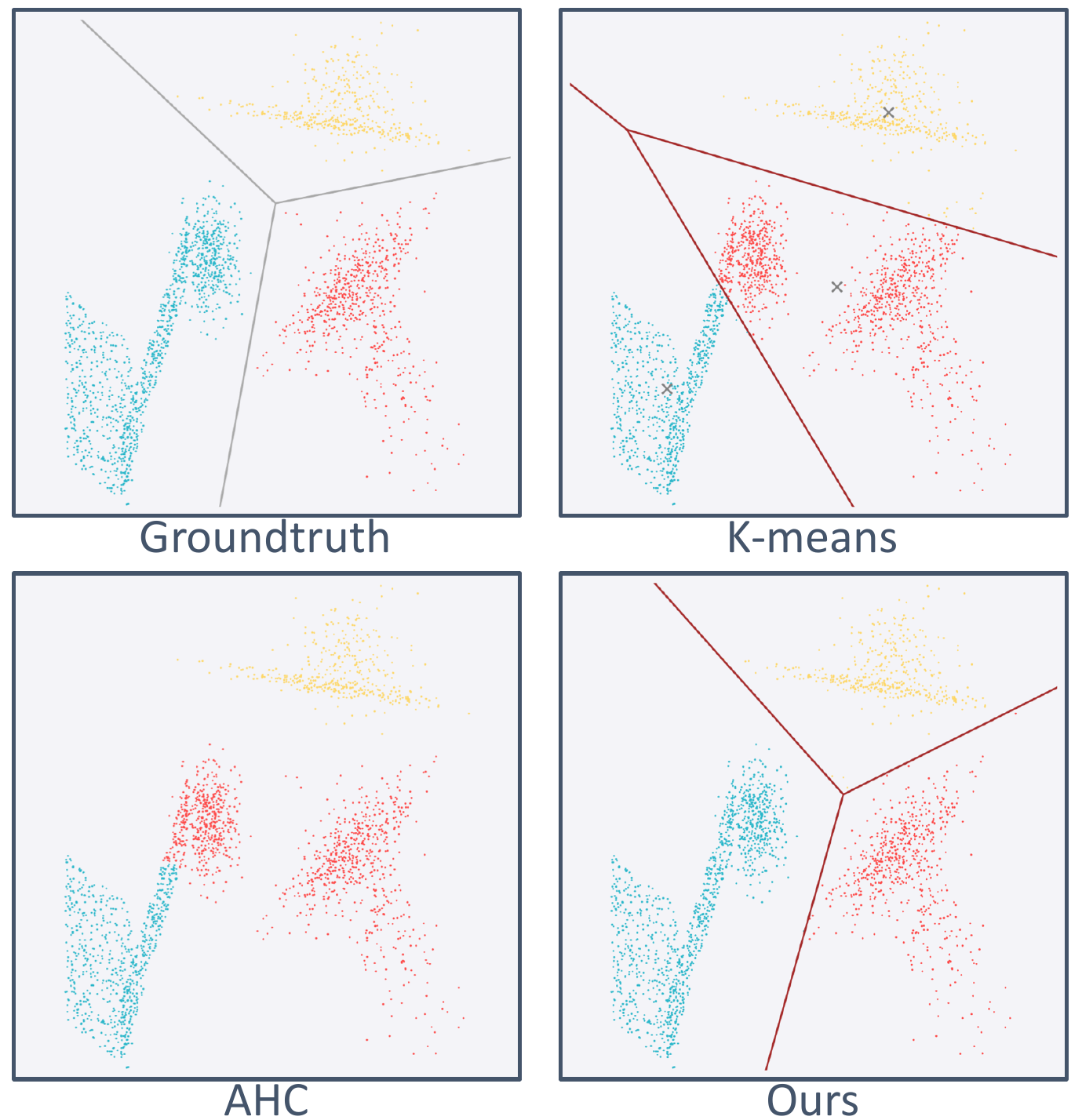}
    \caption{
    An illustration of clustering on linearly separable data.
    Top-left, the groundtruth cluster assignments, and the decision boundaries of the linear SVM trained on the groundtruth.
    Other panels, results of different clustering algorithms.
    Decision boundaries are visualized as solid lines, except for AHC which has no such boundary.
    ``x'' denotes the centroids of K-means.}
    \label{fig:ill_klish}
\end{figure}

\begin{table}[t]
    \centering
    \begin{tabular}{c|ccc}
    \whline{1.0pt}
    GAN & StyleGAN2 & StyleGAN & PGGAN \\
    Dataset & FFHQ & CelebAHQ & CelebAHQ \\\hline
    Linear Classifier \cite{xu2021linear} & 82.7 & 73.8 & 69.3 \\\hline
    Centroids of Features & 49.6 & 46.6 & 41.8 \\
    \whline{1.0pt}
    \end{tabular}
    \caption{The semantics segmentation performance (measured in mIoU) of linear classifier or centroids of features.
    Please see \cref{subsec:ovr_expr} for details.}
    \label{tab:face_iou_centroid}
\end{table}

\label{subsec:centroid_assumption}
The reason might be that K-means and AHC require data to be centered around their class centroids, which is not well satisfied in GAN's features.
To illustrate this, we first took a pretrained GAN and segmented the generated images with a pretrained DeepLabV3 \cite{chen2017rethinking} model.
The features from the GAN were averaged into a centroid for each semantic class.
Then, we segmented another set of generated images by assigning the closest centroid to GAN's features on each spatial location.
It was observed that the segmentation results were rather coarse (\cref{fig:seg_centroid}).

Interestingly, a previous work \cite{xu2021linear} found that GAN's features of a semantic class were linearly separable from other classes.
This observation is referred to as the GAN's linear separability in semantics.
We also found that the linear classifier produced better segmentation than using the centroid method (\cref{tab:face_iou_centroid}).
Therefore, we propose to cluster GAN's features by leveraging their linear separability property, instead of the conventional centering assumption.

We note that linearly separable data does not necessarily satisfy the centering assumption required by conventional clustering algorithms.
To illustrate this, we generated three sets of points in a 2D plane, which were linearly separable from each other (\cref{fig:ill_klish}, top-left). 
In this synthetic dataset, each cluster had a complex structure and its centroid had limited information about the structure.
It was observed that both AHC and K-means assigned a part of the blue cluster to the red cluster.

We propose K-means with Linear Separability Heuristic (KLiSH) to find semantically meaningful clusters in GAN's features.
In brief, KLiSH starts with a large number of initial K-means clusters and iteratively merges the clusters according to their degree of linear separability.
KLiSH successfully clustered the three sets of points on the toy dataset described above (\cref{fig:ill_klish}, bottom-right). 
We compared KLiSH to several clustering algorithms, AHC, K-means, and K-means based Approximate SPectral clustering (KASP) \cite{yan2009fast} on several GANs and datasets.
Results showed that KLiSH outperformed all baselines in most cases.

One can use KLiSH to obtain fine-grained semantic segmentation for the generated images.
The process is straightforward.
First, sample images from GANs and apply the classifier returned by KLiSH on GAN's feature maps.
Then, train downstream models on the resulting image-segmentation dataset.
We demonstrate two applications in this paper: Unsupervised Fine-Grained Segmentation (UFGS) and Unsupervised Semantic-Conditional Synthesis (USCS).
UFGS seeks to reduce the labeling cost of fine-grained semantic annotation on a new dataset.
USCS aims to offer semantic-controllable image synthesis and editing on datasets without semantic annotation.
Both UFGS and USCS can be realized easily with the semantic knowledge extracted from pretrained GANs.

\begin{figure*}[t]
    \begin{center}
        \includegraphics[width=\textwidth]{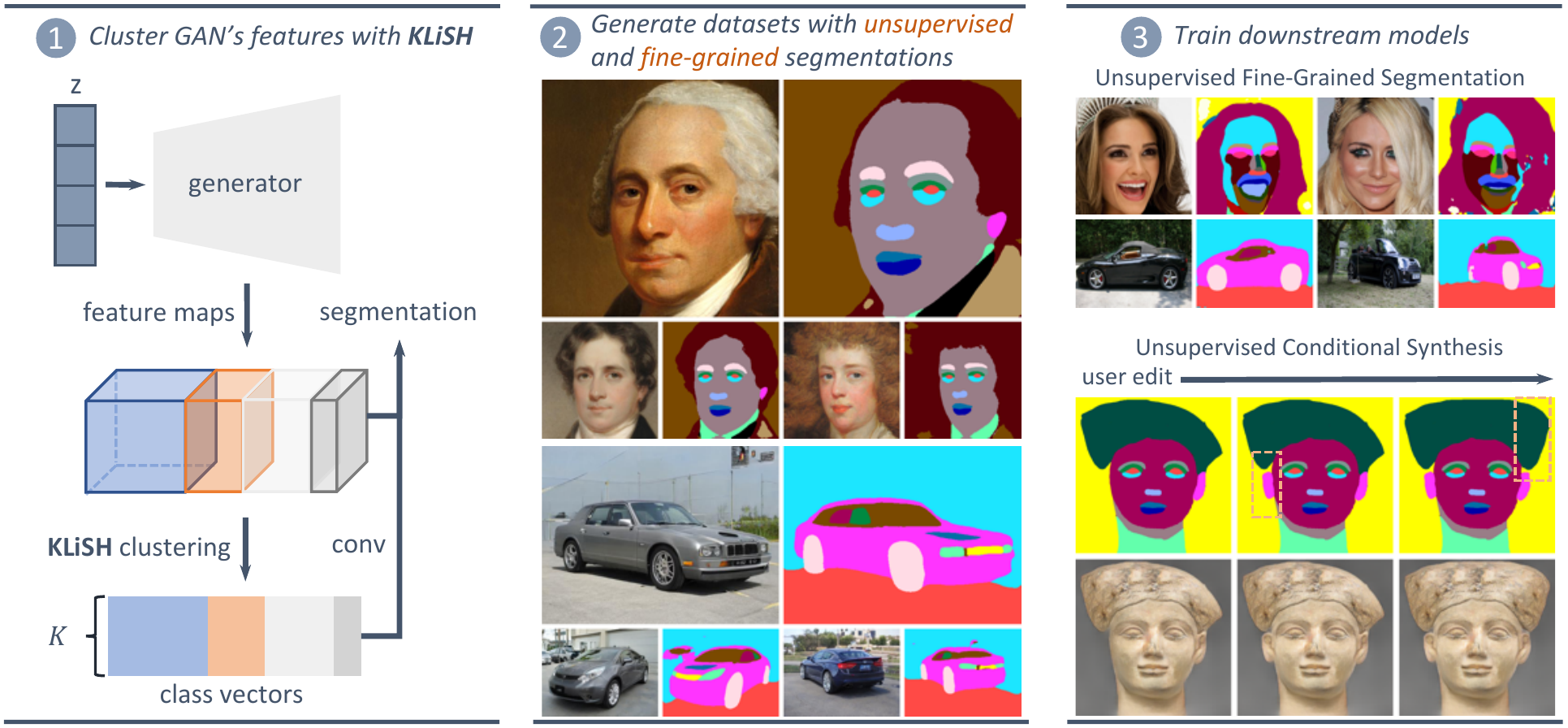}
        \caption{The pipeline of re-purposing pretrained GANs for downstream tasks, UFGS and USCS. First, we propose the K-means with Linear Separability Heuristic (KLiSH) to cluster GAN's features. Then, we generate datasets with synthesized images and segmentation masks together from GANs. Finally, we train different models on the synthesized datasets and perform semantic segmentation on \emph{real} images and semantic-conditional image synthesis. Both applications require no human annotations at all.}
        \label{fig:pipeline}
    \end{center}
\end{figure*}

The pipeline of UFGS and USCS is illustrated in \cref{fig:pipeline}.
Given a dataset of images, we first train a GAN on it and use KLiSH to parse the semantic clusters learned by GANs.
Then, we generate a paired image-segmentation dataset by sampling images from GANs along with the segmentation masks.
Finally, we train corresponding downstream models on the generated dataset.

In summary, our contributions are as follows.
First, we propose KLiSH and improve the quality of clustering on GAN's features.
Second, we provide evidence for GAN's linear separability in semantics.
Third, we propose two novel applications, UFGS and USCS, marking a successful attempt at fine-grained semantic segmentation and semantic-conditional synthesis in an unsupervised learning setting.


%% file: related.tex
\section{Related work}\label{sec:related_work}

\paragraph{Generative Adversarial Networks}
Originally put forward by Goodfellow \etal \cite{goodfellow2014generative}, GANs go through many improvements in training \cite{karras2017progressive,heusel2017gans}, architecture \cite{denton2015deep,odena2017conditional,huang2017stacked,shaham2019singan} and objective \cite{arjovsky2017wasserstein,gulrajani2017improved,kodali2017convergence,chen2016infogan}.
Nowadays, GANs can map latent vectors to diverse and photo-realistic images on various datasets.
Typical data domains are face \cite{radford2015unsupervised,zhao2017energy,berthelot2017began}, car \cite{karras2019style,karras2020analyzing,karras2017progressive}, animal \cite{karras2020training,karras2021alias}, ImageNet \cite{zhang2019self,brock2018large}, \etc.
Among various GANs, StyleGAN2 \cite{karras2020analyzing} is one of the state-of-the-art models.
Another genre of GANs translates images to images \cite{choi2018stargan,choi2020stargan,Zhu2017Unpaired}.
When the input image is a semantic mask, the task is termed Semantic-Conditional Synthesis (SCS).
Typical models capable of SCS include pix2pix \cite{Isola2017pix}, pix2pixHD \cite{wang2018high}, GauGAN \cite{park2019gaugan} and so on \cite{zhu2020sean,lee2020maskgan,jo2019sc,shi2022semanticstylegan}.
Most of them heavily rely on large-scale semantically-annotated datasets.
Sun \etal \cite{sun2021layout} propose to synthesize objects in an image from bounding boxes, which lowers the label requirement, but still needs a large-scale dataset.
In contrast, the USCS proposed in this work is fully unsupervised -- it is realized by making use of the representations learned by pretrained GANs.

\paragraph{GAN interpretation}
Increasing attention is placed on studying GAN's representation.
Most existing works focus on studying GAN's latent vectors and feature maps.
For the first aspect, prior works \cite{shen2020interfacegan,yang2021semantic} find that there are linear boundaries in latent space that separate the positive and negative attributes of the sampled images.
Some other works focus on the unsupervised discovery of those boundaries \cite{jahanian2019steerability,shen2021closed,voynov2020unsupervised}.
For the second aspect, GAN Dissection \cite{bau2018gan} identifies convolution units that have causal relationships with semantics in the generated images.
Xu and Zheng \cite{xu2021linear} propose and validate that GANs encode high-resolution semantics linearly in feature maps.
Other works study the parameters of GANs \cite{bau2020rewriting,wang2021sketch}.
This work proposes an unsupervised approach to extract learned semantic classes from GANs, providing strong evidence that the GAN learns a naturally clustered and linearly separable semantic representation.

\paragraph{Repurposing GAN for semantic segmentation}
Recently, increasing attention is directed towards semantic segmentation using GAN's representation.
Existing works either fall into few-shot learning or unsupervised learning.
The few-shot learning methods typically apply classifiers on a pretrained GAN's feature maps to get semantic segmentation for the generated image \cite{xu2021linear,galeev2021learning,tritrong2021repurposing,zhang2021datasetgan}.
These methods can segment the images well with a limited number of image annotations (ranging from 1 to 40 depending on the complexity of the target data).

For unsupervised learning, the majority of methods \cite{benny2019onegan,singh2019finegan,abdal2021labels4free,melas2021finding,yang2021unsupervised,voynov2021object} focus on decoupling the foreground and background in GANs without external labeling.
Though the segmentation is accurate, the practical usage is limited as only two clusters are segmented.
Methods on obtaining more clusters mostly involve applying K-means++ \cite{Collins20} or AHC \cite{zheng2021unsupervised,kafri2022stylefusion} directly on GAN's features.
He \etal \cite{he2022interpreting} modify K-means by re-weighting channels.
Different from existing methods, we propose a new clustering algorithm, KLiSH, by leveraging a strong characteristic of GAN's features -- linear separability.
We conducted quantitative evaluations and showed the superiority of KLiSH to previous clustering algorithms on GAN's features.

Aside from GAN, other works \cite{gao2021large,gansbeke2021unsupervised,liu2021unsupervised} also study unsupervised semantic segmentation on large-scale datasets, but mainly focus on object-level semantics instead of part-level semantics and their results are not directly comparable to GAN-based methods.

%% file: cluster.tex
\section{Clustering Generator's Features}\label{sec:cluster}


\subsection{Linear Separability of GAN's Features}

\paragraph{Classical linear separability}
KLiSH is based on GAN's linear semantics \cite{xu2021linear}, thus we first introduce this characteristic to make our paper self-contained.

When the generator synthesizes an image, it also produces a set of feature maps.
A $1 \times 1$ convolutional layer is applied to the upsampled and concatenated feature maps, resulting in a score map of semantic classes for the generated image.
As the extraction process is fully linear, the method is named Linear Semantic Extractor (LSE).
The LSE is compared to two nonlinear semantic extractors (NSEs) with more complicated architectures.
The LSE and NSEs are trained under similar settings and evaluated by the accuracy of the semantic masks they extracted compared to the pretrained segmentation network.
Results show that the relative performance gaps between LSE and NSEs are within $3.5\%$ across various datasets and models.
Therefore, it is shown that linear classifiers suffice to separate GAN's representations of different semantics.
For more detailed settings and results, we refer readers to the original paper \cite{xu2021linear}.

The phenomenon that a linear classifier can classify a given dataset is usually described as linearly separable.
Given a dataset $\mA=\{(\x_i, y_i)\}_{i=1}^N$ where $\x_i \in \bR^{D}$ and $y_i \in \{1,\ldots,M\}$, the classical linear separability states that there exists a set of hyperplanes $\W \in \bR^{M \times D}$ and $\mathbf{b} \in \bR^D$ such that
\begin{equation}\label{eq:classical_ls}
    \forall (\x, y) \in \mA, k \neq y, \W_{y} \x + \mathbf{b}_{y} > \W_k \x + \mathbf{b}_k,
\end{equation}
where $\W_k \in \bR^{1 \times D}$ is a row vector, and $\W_y$ denotes the classifier weight for the ground-truth class of a data point $\x$.

\paragraph{Binary linear separability}\label{subsec:binary_ls_def}
Instead of using classical linear separability, we find that binary linear separability is more useful for building our clustering algorithm, which is defined below.

\begin{definition}\label{def:ovr_ls}
Given a dataset $\mA=\{(\x_i, y_i)\}_{i=1}^N$, we say $\mA$ satisfies the binary linear separability if there exists $\W \in \bR^{M \times D}$ and $\mathbf{b} \in \bR^D$ such that $\forall (\x_i, y_i) \in \mA$, $k \in \{1, \dots, M\}$,
\begin{equation}\label{eq:ovr_ls}
    \mathcal{I}(y_i = k) \cdot ( \W_k \x_i + \mathbf{b}_k ) > 1,
\end{equation}
where $\mathcal{I}(y_i = k)$ returns 1 if $y_i = k$ and -1 otherwise.
\end{definition}
In other words, if for each class in the dataset, there is a two-class SVM separating itself from others, then the dataset satisfies the binary linearly separability.
Unless otherwise specified, linear separability refers to \cref{def:ovr_ls} throughout the rest of the paper.

Clearly, \cref{eq:ovr_ls} is harder to satisfy than \cref{eq:classical_ls}.
Clusters separable with a classical linear classifier might not be separable with binary SVMs.
Do features in GAN satisfy the binary linear separability?
We empirically found that the answer is yes.
The experiment is presented in \cref{subsec:ovr_expr}.

\subsection{Clustering with Linear Separability Heuristic}

Now, we describe the algorithm for clustering GAN's features by linear separability.

\label{subsec:init_kmeans}
\paragraph{Initialize from K-means}
We choose K-means for three reasons.
First, K-means is easily scalable to large datasets with time complexity of $O(N)$.
Second, the decision boundaries between clusters are linear and thus are suitable to implement linear separability.
Third, K-means with a large number of clusters provides a good initialization for our algorithm.
More specifically, most clusters found by K-means do not span across multiple semantic classes, see the K-means@100 column in \cref{fig:klish_results}.
This simplifies our problem as we only need to merge these clusters to recover the ground-truth classes.



\paragraph{Merge clusters}\label{subsec:qlin}
Ideally, the merged clusters should belong to the same semantic class.
However, we have no access to semantic information as the algorithm is fully unsupervised.
We make a reasonable assumption: the more linearly separable a cluster is, the more likely it is a true semantic class.
Under this assumption, our goal is to maximize the linear separability of clusters.

How to quantify the degree of linear separability?
According to \cref{def:ovr_ls}, it is implied that how accurately the SVM classifies a cluster indicates how well its linear separability is.
Therefore, we use the performance of the SVM trained on the target dataset to indicate its linear separability.

First, a pseudo-labeled dataset $\mA$ with cluster assignments is constructed.
Second, binary SVMs are trained on $\mA$ using the L2-regularized L2-loss Support Vector Classification objective \cite{fan2008liblinear}:
\begin{multline}\label{eq:svm_loss}
    \mathcal{L} = \frac{\lambda_1}{M |\mA|} \sum_{(\x, y) \in \mA} \bigg [ (1 - \W_y \x - \mathbf{b}_y)_+^2 \\
    + \sum_{k \neq y} (1 + \W_k \x + \mathbf{b}_k)_+^2 \bigg ]
    + \frac{1}{2 M} \norm{\W}_F^2,
\end{multline}
where $(\cdot)_+$ denotes $max(0, \cdot)$ and $\norm{\cdot}_F$ denotes the Frobenius norm.
We use L-BFGS \cite{liu1989limited} to minimize $\mathcal{L}$ due to the convexity of the objective.
The SVM is considered as converged if its $L_{\infty}$ norm of weight changing is below $10^{-4}$.
Third, as in semantic segmentation, we measure the performance of SVM by IoU (Intersection-over-Union) defined as
\begin{equation}\label{eq:iou}
    \displaystyle
    \text{IoU}_k(\W, \mA) = \frac{|\mathcal{S}_k \cap \mathcal{Y}_k|}{|\mathcal{S}_k \cup \mathcal{Y}_k|},
\end{equation}
where $\mathcal{S}_k = \{\x_i | \W_k \x_i + \mathbf{b}_k > 0\}$ is the prediction given by the SVM, $\mathcal{Y}_k = \{\x_i | y_i = k\}$ is the set of positive samples for class $k$, and $(\x_i, y_i) \in \mA$.
Therefore, we can use IoU to find the cluster with the lowest linear separability.

Suppose that the identified cluster is $p$.
The next step is to find another cluster $q$ to merge with $p$.
As implied by \cref{eq:iou}, the cluster $p$ is not classified accurately by the SVM, \ie, the SVM might confuse $p$ with other clusters.
Then a reasonable strategy is to merge $p$ with another cluster that confuses it the most.
To quantify the degree of confusion, we propose the Effective Cosine Similarity (ECoS) metric as defined below.
\begin{definition}
Consider a dataset $\mA=\{(\x_i, y_i)\}_{i=1}^N$.
Suppose a set of binary SVMs with weight $\W \in \bR^{M \times D}$ and bias $\mathbf{b} \in \bR^D$ are trained on $\mA$ until convergence.
The ECoS between cluster $i$ and cluster $j$ is defined as
\begin{equation}\label{eq:ECoS}
\begin{split}
    & \text{ECoS}_{i,j}(\W, \mA) = \cos \langle \mathbf{s}_i^\dagger, \mathbf{s}_j^\dagger \rangle \\
    & \mathbf{s}_k^\dagger = \text{CLAMP}_{[0,1]} \left ( \frac{\mathbf{s}_k + 1}{2} \right ) \\
    & \mathbf{s}_k = \left [ \x_1^T \W_k + \mathbf{b}_k, \ldots, \x_N^T \W_k + \mathbf{b}_k \right ] 
\end{split}
\end{equation}
where $\text{CLAMP}_{[a,b]}$ denotes clamping the input within $[a,b]$.
\end{definition}
Here, $\mathbf{s}_k$ is the cluster scores predicted by the SVM.
$\mathbf{s}_k^\dagger$ scales and truncates $\mathbf{s}_k$ within [0, 1], representing the confidence of SVM predicting positive data points.
As each feature vector corresponds to a pixel, $\mathbf{s}_k^\dagger$ can be resized into a cluster confidence map.
In other words, ECoS measures the cosine similarity between two cluster confidence maps.

\paragraph{Filter initial clusters}
We observed in experiments that a small portion of initial clusters did span across multiple semantic classes.
For example, see K-means@100 of StyleGAN2-FFHQ in \cref{fig:klish_results}, the boundary around ear (visualized as a yellow region) contains part of ``ear'', ``hair'', and ``face''.
This type of initial clusters would harm the accuracy of merged clusters.
Fortunately, we empirically found that these clusters were not linearly separable, \ie, could be filtered by setting a threshold of IoU.
The threshold could also be determined adaptively as described below.

First, train SVMs on the initial clusters and calculate their IoU, resulting in a vector $\mathbf{m}$.
Second, $\mathbf{m}$ is converted from [0, 1] to $(-\infty, \infty)$ by using the inverse sigmoid function $h^{-1}(\cdot)$.
Third, any clusters below $\mu_{\mathbf{m}} - \sigma_{\mathbf{m}}$ are filtered, where $\mu$, $\sigma$ are the mean and standard deviation of $h^{-1}(\mathbf{m})$, respectively. 
Finally, K-means is run again with the filtered centroids as initialization.

\begin{algorithm}[b]
    \KwIn{$\mathcal{D}=\{\x_i\}_{i=1}^N$; $K_0$}
    \KwOut{$\Theta = \{(\W^t, \mathbf{b}^t)\}_{t=1}^{K_0-1}$}
    $\W^0, \mathbf{b}^0, \mA^0$ = K-means($\mathcal{D}, K_0$) \\
    \tcp*[h]{\small Filter initial centriods} \\
    $\W', \mathbf{b}'$ = TrainSVM($\W^0, \mathbf{b}^0, \mA^0$) \\
    $\mathbf{m} = \left [ h^{-1}(\text{IoU}_p (\W', \mathbf{b}', \mA^0)) \right ]_{p=1}^{K_0}$ \\
    $\W^0, \mathbf{b}^0, \mA^0 \leftarrow \text{K-means}(\mathcal{D}, \{\W_p^0 | \mathbf{m}_p > \mu_{\mathbf{m}} - \sigma_{\mathbf{m}}\})$ \\
    \For{$t=1, \ldots, K_0 - 1$}{
        $\W^t, \mathbf{b}^t$ = TrainSVM($\W^{t-1}, \mathbf{b}^{t-1}, \mA^{t-1}$) \\
        $\Theta \leftarrow \Theta \cup \{(\W^t, \mathbf{b}^t)\}$ \\
        $p^t, \phi^t = \min_p \text{IoU}_p (\W^t, \mA^{t-1})$ \\
        $q^t, \psi^t = \max_q \text{ECoS}_{p^t,q} (\W^t, \mA^{t-1})$ \\
        \tcp*[h]{\small Merge cluster $p^t$ into $q^t$} \\
        $\mA^t = \{(\x_i, q^t) | y_i = p^t\} \cup \{(\x_i, y_i) | y_i \neq p^t\}$ \\
        $\W^t \leftarrow \text{cat}(\ldots, \W_{p^t-1}^t, \W_{p^t+1}^t, \ldots)$ \\
        $\mathbf{b}^t \leftarrow \text{cat}(\ldots, \mathbf{b}_{p^t-1}^t, \mathbf{b}_{p^t+1}^t, \ldots)$ \\
    }
    \caption{The KLiSH clustering algorithm.}
    \label{alg:klish}
\end{algorithm}

\paragraph{KLiSH algorithm}
The full KLiSH algorithm is presented in \cref{alg:klish}, where $\mathcal{D}$ is the dataset of GAN's features, $K_0$ is the initial cluster number, and the output $\Theta=\{(\W^t, \mathbf{b}, \phi^t)\}_{t=1}^{K_0-1}$ records the SVM and IoU at each merging step.

The initialization stage has two steps.
First, K-means is applied to the datasets.
Second, filtering is performed as described above, resulting in the initial cluster assignments $\mA^0$.
Then, clusters are iteratively merged for $K_0 - 1$ steps.
In step $t$, an SVM is first trained on dataset $\mA^{t-1}$ , which is obtained from step $t - 1$.
Next, the IoU and ECoS are calculated as defined in \cref{eq:iou} and \cref{eq:ECoS}.
The class with the lowest IoU, denoted by $p^t$, will be merged with the class with the highest ECoS with it, denoted by $q^t$.
The process is repeated until only one class is left.
Without additional information, KLiSH cannot identify a proper number of classes, just like K-means.
We can either slide through cluster numbers and select manually or set up a threshold $\phi^*$ and stop when the IoU reaches the threshold.

%% file: expr.tex
\section{Evaluation of KLiSH}\label{sec:klish_eval}

\subsection{Experiment Setup}\label{subsec:setup}

\paragraph{Pretrained GANs}\label{subsec:pretrained_GANs}
We experimented on four widely used GANs, PGGAN \cite{karras2017progressive}, StyleGAN \cite{karras2019style}, StyleGAN2 \cite{karras2020analyzing}, and StyleGAN2-ADA \cite{karras2020training}.
These GANs were pretrained on various datasets, including FFHQ \cite{karras2020analyzing}, CelebAHQ \cite{liu2015faceattributes}, AFHQ \cite{karras2020training}, MetFace \cite{karras2020training}, and LSUN \cite{yu2015lsun} Car, Bedroom and Church split.
The pretrained models were obtained from the official release \footnote{\url{https://github.com/NVlabs/stylegan3}}.

\begin{table}[t]
    \centering
    \begin{tabular}{lll}
    \whline{1.0pt}
    GANs                            & layer indices     & channels \\\hline
    StyleGAN2-Bedroom              & \multirow{2}{*}{9, 11, 13}         & \multirow{2}{*}{896} \\
    StyleGAN2-Church               &                                    &                      \\\hline
    PGGAN-{[}Bedroom, Church{]}     & \multirow{2}{*}{7, 9, 11, 13}      & \multirow{2}{*}{960} \\
    StyleGAN-{[}Bedroom, Church{]}  &                                    &                      \\\hline
    StyleGAN2-Car                   & \multirow{2}{*}{9, 11, 13, 15}     & \multirow{2}{*}{960} \\
    ADA-{[}Cat, Dog, Wildlife{]}    &                                    &                      \\\hline
    PGGAN-CelebAHQ                  & \multirow{2}{*}{9, 11, 13, 15, 17} & \multirow{2}{*}{496} \\
    StyleGAN-CelebAHQ               &                                    &                      \\\hline
    StyleGAN2-FFHQ                  & \multirow{2}{*}{9, 11, 13, 15, 17} & \multirow{2}{*}{992} \\
    ADA-MetFace                     &                                    &                     \\
    \whline{1.0pt}
    \end{tabular}
    \caption{The indices of GAN layers selected for clustering.}
    \label{tab:selected_layers}
\end{table}

\paragraph{GAN feature selection}\label{subsec:feature_selection}
We only collected features from a subset of GAN's layers mainly due to GPU memory constraints.
To select the layers, we scanned through a generator's layers from high resolution to low and stopped adding new layers if the accumulated number of channels exceeded 1,000.
If multiple layers were having the same resolution, only the last layer of that resolution was selected.
The selected layers for all GANs are shown in \cref{tab:selected_layers}.
The feature maps from the selected layers are then bilinearly interpolated into the same resolution and concatenated along the channel axis.
For clustering, we use $256 \times 256$ resolution.
For generating the segmentation (in UFGS and USCS), we use the full image resolution.

\paragraph{Semantic segmentation networks}\label{sec:segnet_setup}
For face images (CelebAHQ and FFHQ), we trained a DeepLabV3 model with ResNet50 \cite{he2016deep} backbone on CelebAMask-HQ \cite{lee2020maskgan}.
Following \cite{xu2021linear}, we merged duplicate classes like ``left eye'' and ``right eye'', resulting in 15 classes.
The model was trained on the training split of CelebAMask-HQ for 20 epochs.
We used Adam with learning rate $10^{-3}$, $\beta_1=0.9$, $\beta_2=0.999$ and batch size 32.
The learning rate was lowered to $10^{-4}$ after 10 epochs.
We found that 20 epochs were sufficient for the convergence of the DeepLabV3 model.
For bedroom and church images, we used a DeepLabV3 model \footnote{\url{https://github.com/zhanghang1989/PyTorch-Encoding}} pretrained on ADE20K \cite{2017Scene}.

\paragraph{Baseline clustering algorithms}
We compared KLiSH with K-means, AHC, and KASP \cite{yan2009fast}.
K-means and AHC have been used to cluster GAN's features \cite{Collins20,he2022interpreting,zheng2021unsupervised,kafri2022stylefusion}, but they implicitly require the centroid assumption (described in \cref{subsec:centroid_assumption}).
Spectral clustering does not have such a constraint, however, it is not practical for our task as it has an $O(N^3)$ time complexity \cite{yan2009fast}.
Therefore, we consider using KASP, a fast approximate algorithm for spectral clustering.
Interestingly, KASP also performs K-means first and then merges the initial clusters.
Different from KLiSH, KASP uses spectral clustering on the initial centroids to merge them and does not use linear separability.

We implemented a multi-GPU version of K-means++.
KASP was implemented following the original paper \cite{yan2009fast}.
For AHC, we made two modifications based on the scikit-learn implementation \footnote{\url{https://scikit-learn.org/stable/modules/generated/sklearn.cluster.AgglomerativeClustering.html}}.
First, we trained a linear classifier on the cluster assignments obtained by AHC for prediction; otherwise, AHC would rebuild the whole merging tree to predict new data, which was impractical.
Second, we gave AHC less number of features due to its expensive memory cost.
More optimization of AHC might be possible, but would be beyond the scope of this work.

\paragraph{Evaluation of clustering}\label{sec:cluster_setup}
For each GAN, we sampled a total of 256 images, resulting in a block of features with shape $(256, 256, 256, D)$.
The only exception is AHC, where we used 16 images and $64 \times 64$ resolution, due to the expensive memory cost mentioned above.
Then, we applied KLiSH, K-means, AHC, and KASP to the sampled features and evaluated them on the same set of 10k sampled images (different from those used for clustering).

For evaluation, we consider two widely used metrics, Adjusted Mutual Information (AMI) and Adjusted Rand Index (ARI).
Both AMI and ARI are designed for general-purpose clustering tasks and may not reflect the performance of segmentation tasks faithfully.
In the context of (supervised) semantic segmentation, mIoU is the dominant metric.
Therefore, we also propose the Maximum matching mIoU (MIoU) metric defined below,
\begin{align}
    \displaystyle
    \text{MIoU}(\{\mathcal{Y}_m\}_{m=1}^M, \{\mathcal{C}_k\}_{k=1}^K) & =  \max_{\mathbf{a} \in \{0, \ldots, M\}^K} \frac{1}{M} J(\mathbf{a}), \\
    J(\mathbf{a}; \{\mathcal{Y}_m\}, \{\mathcal{C}_k\}) & = \sum_{m=1}^M \text{IoU}(\mathcal{Y}_m, \bigcup_{\{k | \mathbf{a}_k = m\}} \mathcal{C}_k),
\end{align}
where $M$ is the number of groundtruth classes, $K$ is the number of clusters, $\mathcal{Y}_m$ is the groundtruth class $m$, and $\mathcal{C}_r$ is the predicted cluster $r$.
$\mathbf{a}$ is a $K$-dimensional vector that encodes the matched class for each cluster.
If the $r$-th element of $\mathbf{a}$, $\mathbf{a}_r$ equals 0, then the cluster $r$ does not match any groundtruth class. 
Otherwise, $\mathbf{a}_r > 0$ indicates cluster $r$ is matched to groundtruth class $\mathbf{a}_r$.
In short, MIoU measures the segmentation performance of the predicted clusters after they have been matched correctly to the groundtruth classes.
MIoU should only be compared when two clustering algorithms produce the same number of clusters, \ie, when the cost of label permutation is identical.
This constraint is also required by other clustering metrics such as AMI and ARI.

\begin{figure}[t]
    \centering
    \includegraphics[width=0.98\linewidth]{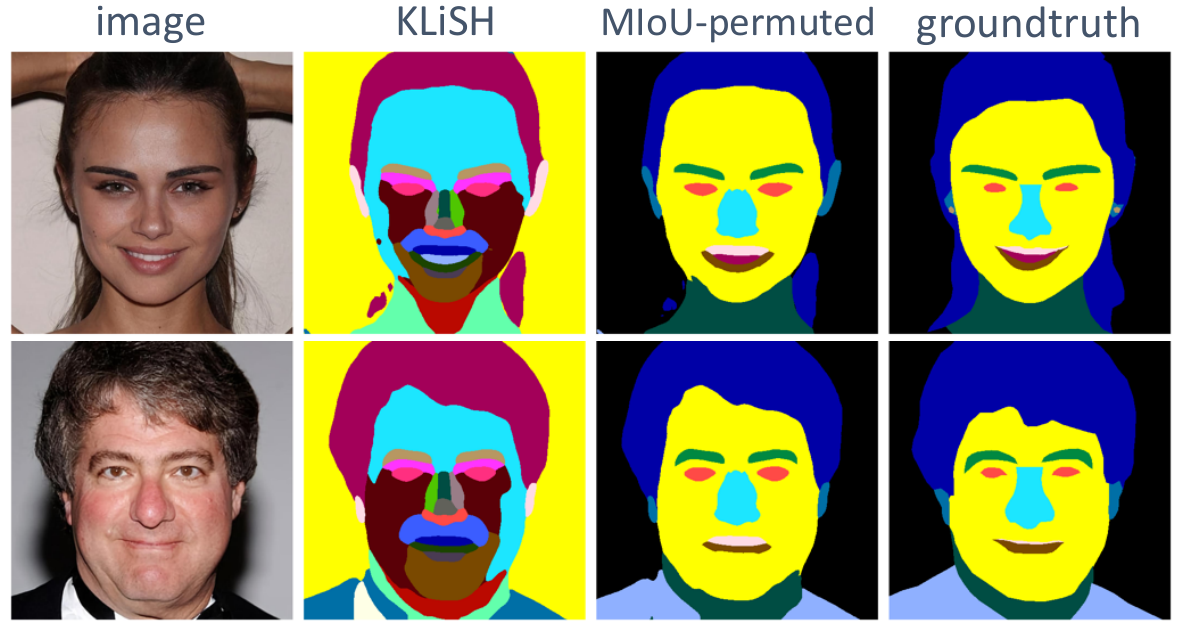}
    \caption{Demonstration of MIoU-permuted labels.}
    \label{fig:MIoU_permute}
\end{figure}

The MIoU is calculated by a greedy algorithm as described in \cref{alg:MIoU}, in which $(\mathbf{a}|_{\mathbf{a}_k \leftarrow m})$ denotes that the $k^{\text{th}}$ element of $\mathbf{a}$ is set to $m$.
To demonstrate the effectiveness of MIoU, we tested \cref{alg:MIoU} on a few images.
As shown in \cref{fig:MIoU_permute}, the label permutations found by \cref{alg:MIoU} well matched the predicted clusters to the groundtruth classes.

\begin{algorithm}[t]
    \KwIn{$\{\mathcal{Y}_m\}_{m=1}^M, \{\mathcal{C}_k\}_{k=1}^K$}
    \KwOut{MIoU, $\mathbf{a}^t \in \{0, 1, \ldots, M\}^K$}
    $\mathbf{a}^0=\mathbf{0}$ \\
    \For{$t=1, \ldots, K$}{
        $\mathcal{I}^t = \{k | \mathbf{a}_k^{t-1} = 0\} \times \{1, \ldots, M\}$ \\
        $\mathcal{G}^t = \{\mathbf{a}' | \mathbf{a}' = (\mathbf{a}^{t-1}|_{\mathbf{a}_k^{t-1} \leftarrow m}), (k, m) \in \mathcal{I}^t\}$ \\
        $\mathbf{a}^t = \max_{\mathbf{a}' \in \mathcal{G}^t} J(\mathbf{a}'; \{\mathcal{Y}_m\}, \{\mathcal{C}_k\})$ \\
    }
    MIoU = $J(\mathbf{a}^t; \{\mathcal{Y}_m\}, \{\mathcal{C}_k\})$
    \caption{Maximum matching mIoU.}
    \label{alg:MIoU}
\end{algorithm}

All experiments were conducted using PyTorch \cite{pytorch2019}.
KLiSH required 8 GeForce RTX 2080 Ti GPUs to run and took less than 30 minutes per GAN model.

\begin{figure*}[p]
    \centering
    \includegraphics[width=\textwidth]{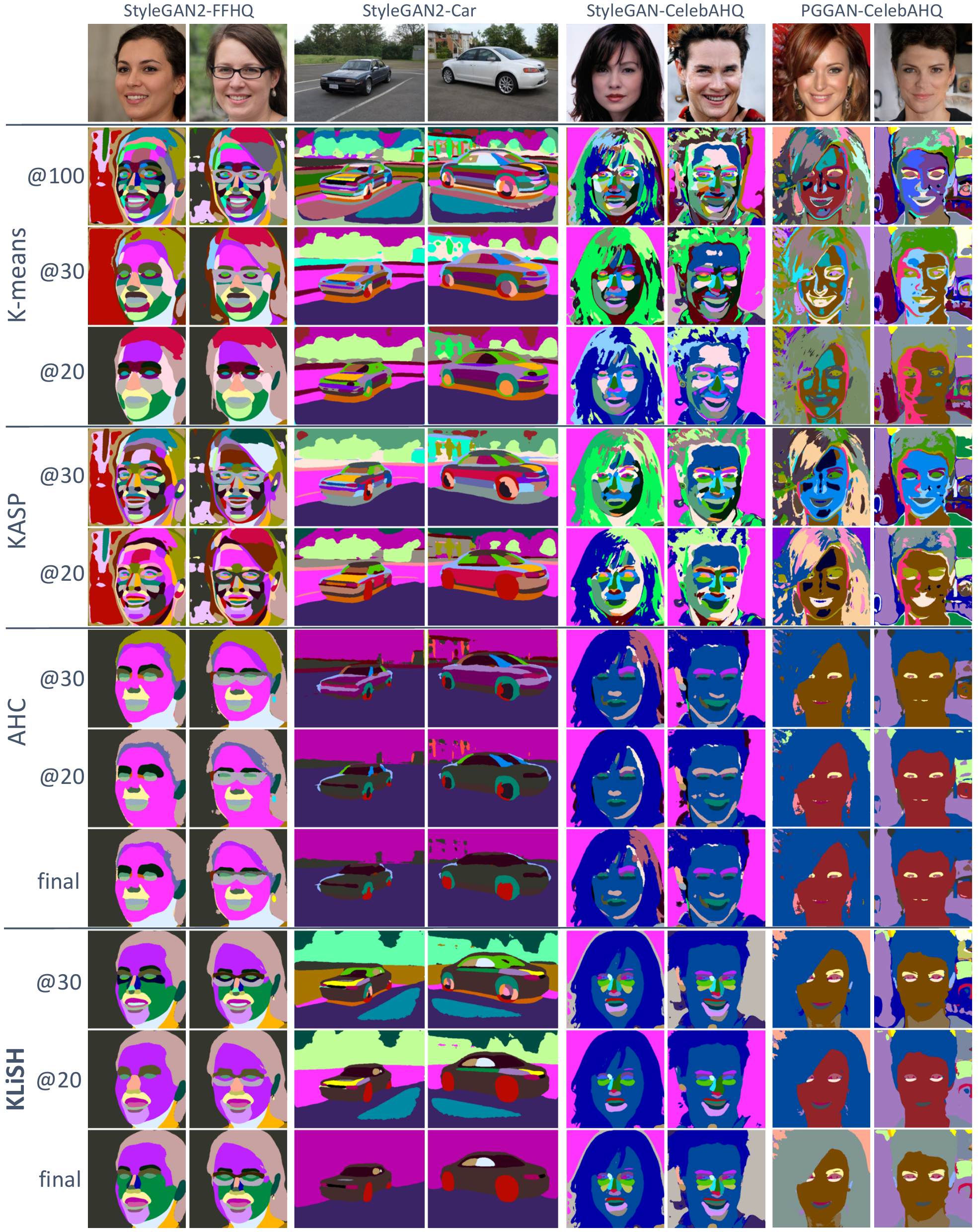}
    \caption{Results of K-means, AHC, KASP, and KLiSH on face images and car images. The first row shows the generated images and the other rows show the clustering results obtained with different cluster numbers, denoted by the number after ``@''.
    ``final'' refers to the selected number of clusters for downstream tasks, which is 26, 11, 30, and 26 from left to right, respectively.
    }
    \label{fig:klish_results}
\end{figure*}

\begin{table*}[t]
    \centering
    \begin{tabular}{c|ccc|ccc|ccc}
    \whline{1.0pt}
    GAN & \multicolumn{3}{c|}{StyleGAN2} & \multicolumn{3}{c|}{StyleGAN} & \multicolumn{3}{c}{PGGAN} \\
    Dataset & FFHQ & Bedroom & Church & CelebAHQ & Bedroom & Church & CelebAHQ & Bedroom & Church \\\hline
    SVM          & 81.7 & 43.5 & 33.7 & 72.2 & 30.9 & 31.4 & 66.4 & 27.7 & 45.4 \\
    LSE         & 82.7 & 45.2 & 35.2 & 73.8 & 37.3 & 34.8 & 69.3 & 31.1 & 47.8 \\\hline
    $\Delta\%$ & -1.3 & -3.8 & -4.3 & -2.2 & -9.8 & -17.2 & -4.2 & -10.9 & -5.0 \\
    \whline{1.0pt}




    \end{tabular}
    \caption{The semantic extraction performance (in mIoU\%) of SVM and LSE on various GANs and datasets. $\Delta$ denotes the relative difference, $\Delta = \frac{\text{SVM} - \text{LSE}}{\text{LSE}}$.}
    \label{tab:face_iou_extractor}
\end{table*}

\subsection{Binary Linear Separability Results}\label{subsec:ovr_expr}
As described in \cref{subsec:binary_ls_def}, KLiSH relies on the binary linear separability of GAN's features (\cref{def:ovr_ls}).
To test if this assumption is satisfied, we compared the performance of semantic segmentation using GAN's features using either the SVM or the LSE.
The LSE shared the same architecture with the SVM but was trained with cross-entropy loss, reflecting the classical linear separability.
For each GAN, we sampled 50k images, collected their features, and obtained their segmentation masks using the pretrained DeepLabV3 model.
Next, we trained LSE and SVM on the features and segmentation masks for 1 epoch using the Adam optimizer with $10^{-3}$ learning rate.
The SVM used \cref{eq:svm_loss} with $\lambda_1=5000$, which was identical to the value used in KLiSH.
For evaluation, we sampled another different 10k images and measured the mIoU between the segmentation of LSE or SVM and the segmentation of the DeepLabV3 model.

Their performance is presented in \cref{tab:face_iou_extractor}.
SVM achieved close performance with LSE in most cases.
The relative differences between SVM and LSE were within 5\% on 6 models.
Therefore, in general, the classical linear separability of GAN's features can be relaxed to the binary linear separability.

\begin{table}
    \centering
    \resizebox*{\linewidth}{!}{
    \begin{tabular}{cc|c|c|cc|c}
    \whline{1.0pt}
    \multirow{2}*{K} & \multirow{2}*{metric} & K-means & KASP & \multicolumn{2}{c|}{AHC} & KLiSH (ours) \\
    & & {\footnotesize euclidean} & {\footnotesize euclidean} & {\footnotesize ward} & {\footnotesize arccos} & {\footnotesize euclidean} \\\hline
\multicolumn{7}{c}{StyleGAN2-FFHQ} \\\hline
\multirow{3}*{20} &  AMI & 64.8 & 61.9 & 62.9 & 65.2 & \textbf{70.6} \\
                  &  ARI & 57.1 & 54.4 & 60.7 & 65.1 & \textbf{70.5} \\
                  & MIoU & 50.2 & 40.3 & 46.4 & 37.7 & \textbf{52.7} \\\hline
\multirow{3}*{$26^\dagger$} &  AMI & 56.3 & 59.1 & 58.0 & 68.0 & \textbf{68.3} \\
                  &  ARI & 25.1 & 37.2 & 37.7 & \textbf{69.2} & 65.4 \\
                  & MIoU & 52.6 & 48.9 & 46.9 & 43.8 & \textbf{56.3} \\\hline
\multirow{3}*{30} &  AMI & 56.3 & 59.3 & 57.7 & 65.9 & \textbf{67.6} \\
                  &  ARI & 25.1 & 38.5 & 35.7 & 64.2 & \textbf{64.6} \\
                  & MIoU & 52.6 & 53.2 & 49.9 & 45.5 & \textbf{58.0} \\\hline
\multicolumn{7}{c}{StyleGAN-CelebAHQ} \\\hline
\multirow{3}*{20} &  AMI & 53.0 & 55.5 & 53.1 & 62.5 & \textbf{69.0} \\
                  &  ARI & 38.3 & 38.1 & 49.1 & 67.0 & \textbf{73.1} \\
                  & MIoU & 30.9 & 36.8 & 30.5 & 25.4 & \textbf{41.5} \\\hline
\multirow{3}*{$30^\dagger$} &  AMI & 51.9 & 49.7 & 43.4 & 57.9 & \textbf{63.6} \\
                  &  ARI & 28.2 & 30.5 & 21.4 & 54.8 & \textbf{61.8} \\
                  & MIoU & 38.5 & 33.5 & 31.7 & 27.7 & \textbf{45.7} \\\hline
                  \multicolumn{7}{c}{PGGAN-CelebAHQ} \\\hline
\multirow{3}*{20} &  AMI & 37.9 & 39.3 & 39.4 & 37.1 & \textbf{51.0} \\
                  &  ARI & 18.5 & 19.4 & 20.5 & 30.1 & \textbf{53.5} \\
                  & MIoU & 15.2 & 16.9 & 16.4 & 19.1 & \textbf{23.1} \\\hline
\multirow{3}*{$26^\dagger$} &  AMI & 35.1 & 37.9 & 37.6 & 36.0 & \textbf{51.0} \\
                  &  ARI & 11.5 & 15.3 & 15.6 & 28.7 & \textbf{53.5} \\
                  & MIoU & 17.4 & 19.0 & 16.4 & 19.2 & \textbf{29.0} \\\hline
\multirow{3}*{30} &  AMI & 35.1 & 36.2 & 38.0 & 38.1 & \textbf{50.2} \\
                  &  ARI & 11.5 & 12.4 & 15.4 & 30.7 & \textbf{52.4} \\
                  & MIoU & 17.4 & 18.6 & 18.4 & 19.6 & \textbf{29.8} \\\hline
     \whline{1.0pt}
    \end{tabular}}
    \caption{Quantitative evaluation of clustering algorithms on face image generators. The final number of clusters used is indicated by a $\dagger$. For all metrics, the larger the better. The clustering results are reported as the best one among the 5 trials.}
    \label{tab:eval_cluster}
\end{table}

\subsection{KLiSH Clustering Results}

\paragraph{Quantitative evaluations}\label{sec:quant_eval_cluster}
We evaluated the performance of clustering algorithms on GANs trained on face images.
We report the metrics at cluster numbers K=20, K=30, and the finally selected number in \cref{tab:eval_cluster}.

It was clear that KLiSH outperformed existing clustering algorithms.
For AMI and MIoU, KLiSH was superior to all baselines in all cases.
Notably, at the finally selected number of clusters, KLiSH achieved significantly higher MIoU scores.
It indicated that KLiSH achieved the best segmentation performance at the cluster number that would be used by downstream tasks.
For ARI, KLiSH also achieved the highest score in almost every case, except for K=26 in StyleGAN2-FFHQ, where AHC (arccos) achieved a better score.
Even though, qualitative results showed that AHC was not better than KLiSH in this case (see row ``final'' of column ``StyleGAN2-FFHQ'' of \cref{fig:klish_results}).
AHC failed to cluster ``eye'', ``nose'', ``upper lip'', ``lower lip'', but KLiSH successfully identified all those clusters.
KLiSH's lower ARI might result from leaving two clusters for ``face'' and ``neck''.
However, in this case, it would be rather easy to merge them manually.
In contrast, dividing the undersegmented clusters obtained by AHC would be more difficult.

In summary, KLiSH was better than K-means, AHC, and KASP on GAN's features.

\paragraph{Qualitative results}\label{sec:qual_eval_cluster}
We visualize the clusters obtained by all the cluster algorithms on various GANs and datasets in \cref{fig:klish_results}.
The results were from the same trial (best out of 5 trials) as in \cref{tab:eval_cluster}.
We had two main observations.


First, KLiSH was significantly more accurate than K-means, AHC, and KASP.
In StyleGAN2-FFHQ, K-means@30 failed to cluster ``upper lip'', ``lower lip'', ``teeth'', ``brow'' classes and still undersegmented ``background'' and ``hair''.
AHC@30 already failed to cluster ``nose'' and the classes in mouth region.
In contrast, KLiSH@30 had no undersegmentation of groundtruth classes.
Similarly, for car images, KLiSH@final successfully segmented the ``headlight'', ``wheel'', and even two ``side glasses''.
In contrast, the most competitive baseline, AHC@final not only undersegmented ``background'' and ``wheel'' classes, but also completely missed the ``headlight''.

\begin{figure*}[t]
    \centering
    \includegraphics[width=\textwidth]{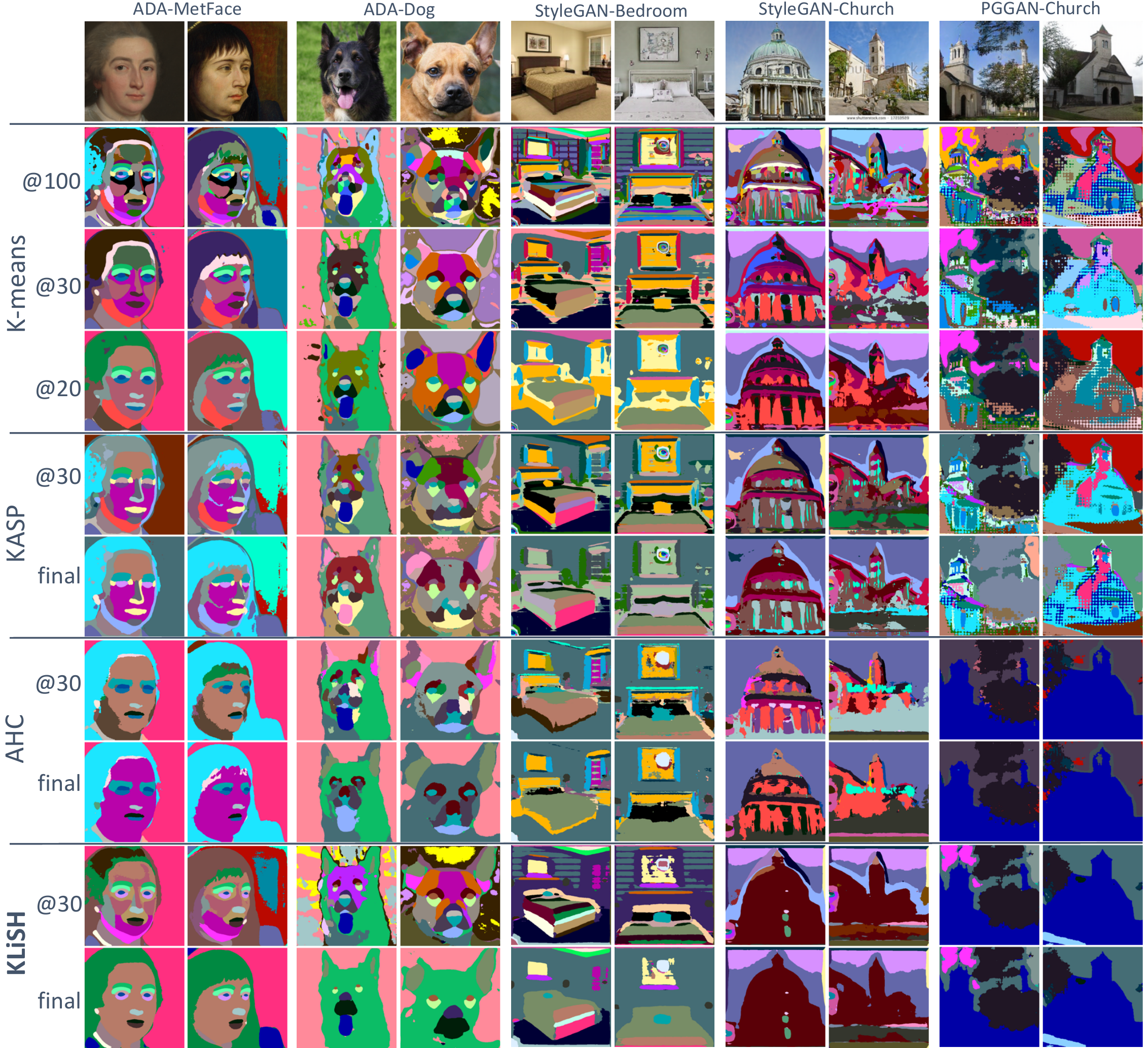}
    \caption{Comparison of clustering algorithms on various datasets and GANs.
    ``final'' refers to 18, 9, 16, 15, and 28 from left to right.
    }
    \label{fig:klish_results_add}
\end{figure*}

Second, the advantage of KLiSH over existing clustering algorithms was consistent across GANs and datasets.
For StyleGAN-CelebAHQ, AHC@final missed ``nose'', ``lip'', and ``teeth'' classes and undersegmented ``hair'', while KLiSH@final did not.
For PGGAN-CelebAHQ, although all the clustering algorithms did not obtain good results -- probably because PGGAN had learned the dataset well.
Still, KLiSH@final successfully segmented ``ear'' and ``hair'' classes, but AHC@final missed ``ear'' and mixed ``hair'' with ``background'' (see the rightmost column).
We provide more visualization in \cref{fig:klish_results_add}, which is also supportive of the observations described above. 

\begin{figure}[t]
    \centering
    \includegraphics[width=0.99\linewidth]{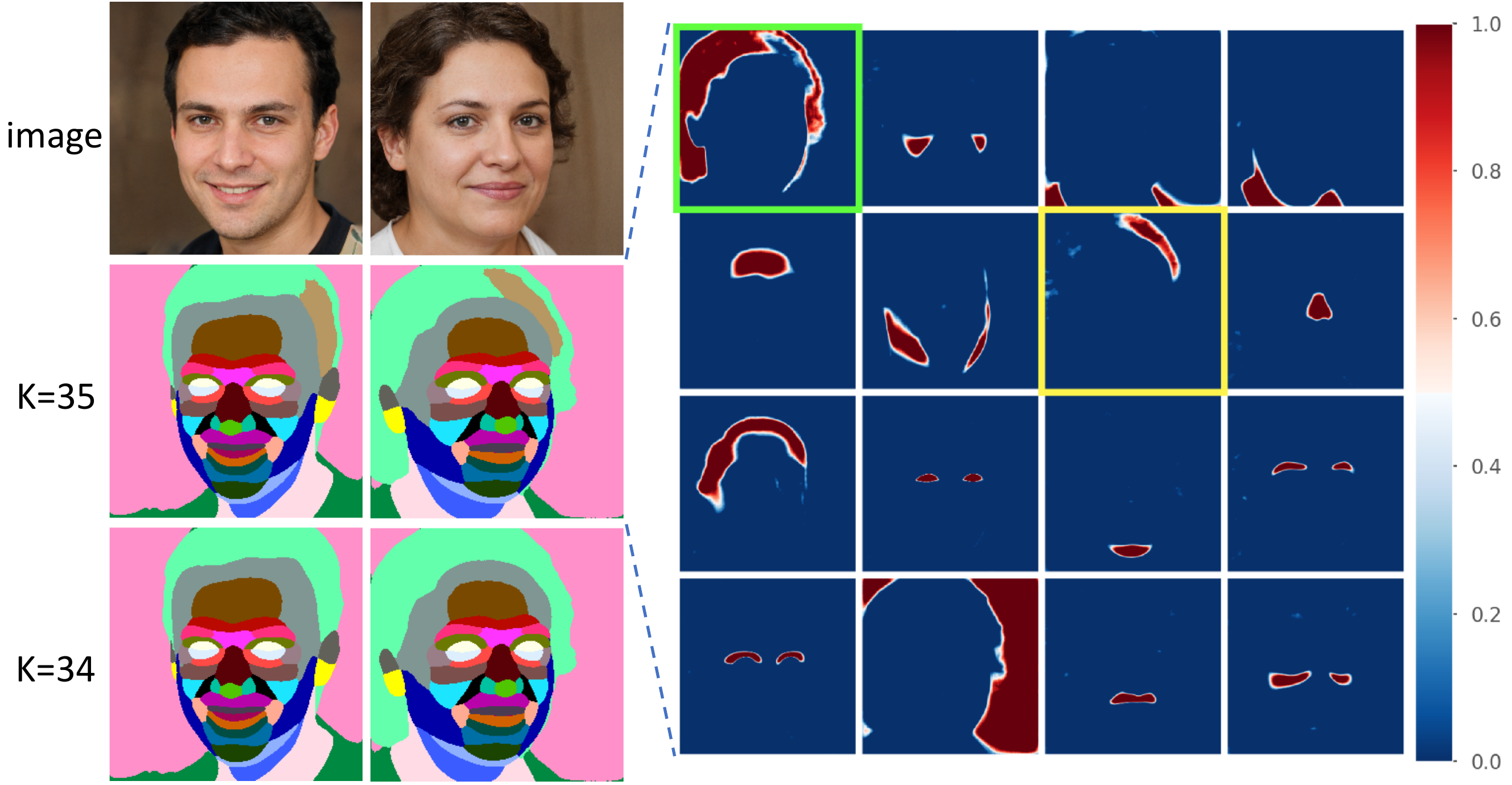}
    \caption{Demonstration of a merging step. The first two columns show the images and clusters.
    The confidence maps $\{\mathbf{s}_k^\dagger\}_{k=1}^M$ of the image are visualized (partially) in other columns.
    $\{\mathbf{s}_k^\dagger\}_{k=1}^M$ take value within [0, 1] and are visualized using the color bar on the right.
    The cluster in yellow box (say cluster $p$) has the smallest IoU (\cref{eq:iou}) with K=35.
    The cluster in green box (say cluster $q$) has the largest ECoS (\cref{eq:ECoS}) with cluster $p$.
    Cluster $p$ and $q$ are then merged in this step.
    }
    \label{fig:merge_demo}
\end{figure}

\paragraph{Visualization of a merging step}
To help readers gain a concrete picture of how KLiSH merged the clusters, we also visualized a single merging step in \cref{fig:merge_demo}.
See the cluster confidence map of ``nose'' and ``brow'' (Rows 2 and 3, Column 4 in the right figure), the red regions are dominated by deep red, indicating that the SVM predicts them with high confidence.
On the contrary, the heatmap denoted by the yellow box has a large proportion of shallow red, indicating that the confidence of SVM to predict that cluster was low.
This cluster had the lowest IoU at K=35 and was subsequently merged with another oversegmented cluster of hair, resulting in better cluster assignments.


%% file: application.tex
\section{Applications}\label{sec:application}

\begin{table}[t]
    \centering
    \begin{tabular}{ccc|c}
    \whline{1.0pt}
    Method & \#shots & \#classes & mIoU(\%) \\\hline
    UFGS (KLiSH) & 0 & 15 & 47.2 \\\hline
    UFGS (AHC) & 0 & 15 & 36.6 \\\hline
    \multirow{2}*{LSE\cite{xu2021linear}} & 1 & \multirow{2}*{15} & 47.1 $\pm$ 2.8 \\
     & 10 & & 58.5 $\pm$ 0.9 \\\hline
    RepurposeGAN \cite{tritrong2021repurposing} & 10 & 9 & 68.0 \\\hline
    DatasetGAN \cite{zhang2021datasetgan} & 16 & 8 & 70.0 \\
    \whline{1.0pt}
    \end{tabular}
    \caption{The performance of two UFGS methods and several few-shot learning methods.}
    \label{tab:face_iou_global}
\end{table}

\begin{table*}[t]
    \centering
    \resizebox*{0.98\textwidth}{!}{
    \begin{tabular}{cc||cccccccccccccc|c}
    \whline{1.0pt}
    {\small method} & \#samples & skin & nose & eye-g & eye & brow & ear & teeth & u-lip & l-lip & hair & hat & ear-r & neck & cloth & mouth* \\
    \hline\hline
    UFGS (KLiSH) & (K=26) & 82.2 & 76.0 & 25.7 & 50.4 & 60.2 & 50.3 & 32.1 & 52.2 & 51.0 & 76.5 & 0.0 & 0.0 & 70.1 & 34.0 & - \\
    UFGS (AHC) & (K=26) & 62.0 & 19.3 & 27.6 & 62.6 & 30.6 & 49.3 & 0.00 & 47.1 & 60.3 & 80.5 & 0.1 & 11.0 & 34.6 & 27.2 & - \\\hline
    \multirow{2}*{\small LSE} & N=1 & 83.8 & 81.7 & 1.9 & 72.9 & 51.0 & 45.6 & 32.7 & 64.8 & 68.2 & 69.0 & 0.6 & 1.7 & 68.8 & 16.1 & - \\
    & N=10 & 89.3 & 85.5 & 25.7 & 72.2 & 61.4 & 58.9 & 60.5 & 70.1 & 74.2 & 85.2 & 6.3 & 11.5 & 74.7 & 43.6 & - \\\hline
    \multirow{2}*{\small RepurposeGAN} & N=1 & - & 76 & - & 58 & - & - & - & - & - & - & - & - & - & - & 71 \\
    & N=10 & 90 & 85 & - & 75 & 68 & 37 & - & - & - & 84 & - & - & 73 & 16 & 87 \\
    \whline{1.0pt}
    \end{tabular}}
    \caption{IoU(\%) per class of DeepLabV3 trained on synthetic datasets generated with different annotation methods.
    The evaluation is conducted on real images from the test split of CelebAMask-HQ.
    ``eye-g'', ``u-lip'', ``l-lip'' denotes ``eyeglasses'', ``upper lip'', and ``lower lip'', respectively.
    The mouth class is a union of ``teeth'', ``u-lip'', and ``l-lip''.
    Our UFGS achieved close performance with one-shot LSE on most classes and outperformed one-shot LSE on {\it hard} classes like ``eyeglasses'' and ``cloth''.}
    \label{tab:face_iou_classwise}
\end{table*}

\begin{figure}[t]
    \centering
    \includegraphics[width=0.99\linewidth]{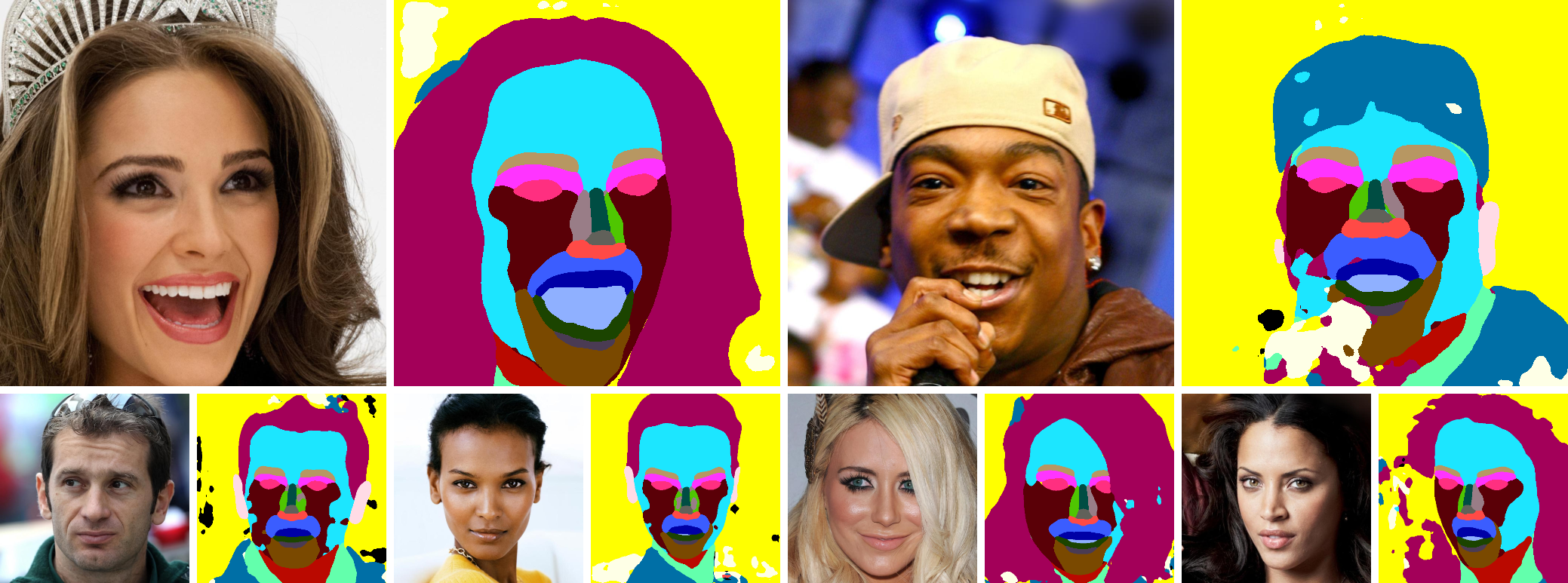}
    \\
    ~\vspace{-3mm}
    \\
    \includegraphics[width=0.99\linewidth]{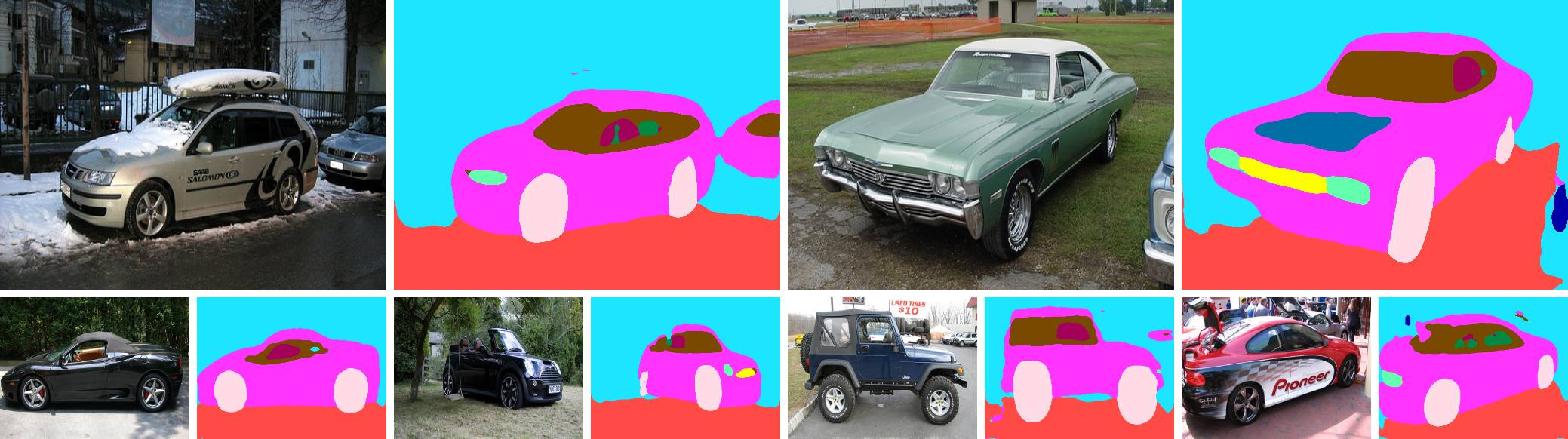}
    \caption{UFGS results on \emph{real} images from CelebAMask-HQ (rows 1 and 2) and PASCAL VOC (rows 3 and 4).}
    \label{fig:UFGS}
\end{figure}

\begin{figure*}[t]
    \centering
    \includegraphics[width=0.98\linewidth]{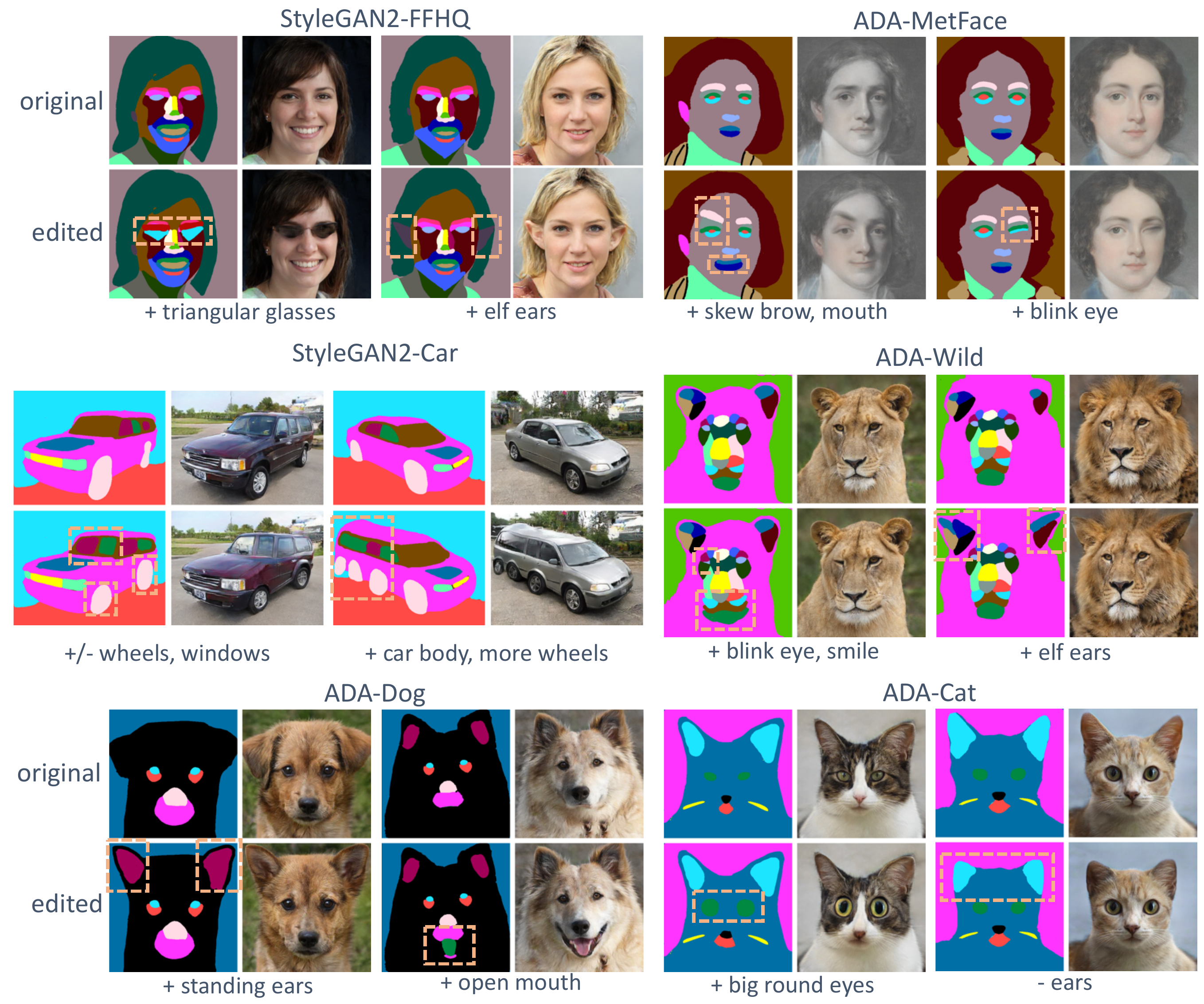}
    \caption{The qualitative results of USCS on various GANs and datasets.
    The semantic masks are drawn by users and the modifications on the masks are indicated by dashed boxes.
    The image on the right of each mask is generated by our USCS model.
    }
    \label{fig:uscs}
\end{figure*}


\subsection{Unsupervised Fine-Grained Segmentation}\label{sec:UFGS}

Semantic segmentation labeling is very expensive.
As described by Lee \etal in CelebAMask-HQ \cite{lee2020maskgan}, annotating fine-grained semantic segmentation is not only time-consuming but also involves multiple iterations of refinement.
To reduce the labeling cost, we propose an approach for Unsupervised Fine-grained Segmentation (UFGS).

Existing methods on unsupervised segmentation \cite{benny2019onegan,abdal2021labels4free,melas2021finding,yang2021unsupervised} only segment foreground from background, thus are not applicable to UFGS.
A more closely related task to UFGS is few-shot learning-based segmentation \cite{xu2021linear,tritrong2021repurposing,zhang2021datasetgan}.
Although they reduce the number of annotations, the cost of fine-grained semantic annotations is still significant.
Zhang \etal \cite{zhang2021datasetgan} reports that annotating a single face image cost on average 20 minutes \emph{per image}.
In contrast, the only labeling cost involved in UFGS is naming and merging a few labels, which could take less than one minute for a \emph{whole dataset}.

\paragraph{Method}\label{subsec:ufgs_mtd}
The proposed method for UFGS is as follows:
First, train a GAN on the target dataset, run KLiSH on the GAN, and select a proper number of semantic clusters.
Next, generate an image-segmentation dataset by sampling images from the GAN and using the SVM returned by KLiSH to obtain the segmentation masks of the image.
We apply the SVM directly on the feature maps upsampled to the resolution of images.
Then, train a segmentation network, \eg, DeepLabV3 \cite{chen2017rethinking}, on the synthetic dataset.
Finally, segment \emph{real} images in the original dataset.

\paragraph{Evaluation}\label{subsec:ufgs_eval}
We evaluated the performance of our UFGS method on face images using the test split of the CelebAMask-HQ dataset.
The cluster assignments were matched to the ground-truth classes by maximizing MIoU (\cref{alg:MIoU}) using 100 images from the validation split.
Note that the images and labels are not used to train the model.
In practice, users can match clusters to labels with just several clicks.
The performance of our method was compared to few-shot learning methods \cite{xu2021linear,tritrong2021repurposing,zhang2021datasetgan}.

To the best of our knowledge, except for face images, there are no large-scale semantic annotations on the datasets described in \cref{subsec:setup}.
PASCAL VOC \cite{PASCAL2010} contains car images, but they have a significant domain gap with the car images from the GAN's dataset, LSUN-Car.
In two relevant works, DatasetGAN\cite{zhang2021datasetgan} and RepurposeGAN \cite{tritrong2021repurposing}, the authors make their own test sets by filtering PASCAL VOC and have not released them as of now.
Therefore, we only present qualitative results on PASCAL VOC and report quantitative results on CelebAMask-HQ.

\paragraph{Experiment setup}
We used StyleGAN2-FFHQ and StyleGAN2-Car for UFGS on CelebAMask-HQ and PASCAL VOC cars, respectively.
The baseline for comparison was UFGS with AHC, as it was the most competitive method with KLiSH on StyleGAN2-FFHQ according to \cref{tab:eval_cluster}.
The settings were identical to UFGS with KLiSH except that the datasets were generated with clusters found by AHC.
We also compared UFGS with KLiSH to a few-shot semantic extractor, LSE.
We reproduced LSE on StyleGAN2-FFHQ and StyleGAN2-Car following the settings in \cite{xu2021linear}.
In brief, we randomly sampled 1 or 10 images (for one-shot or 10-shot experiments) and obtained their segmentation using the pretrained DeepLabV3 model.
Then, we trained the LSE on the features and labels of those images for 6,400 iterations.
To address the variation of few-shot experiments, we trained LSE five times, each time with a different set of image annotations.

After training the LSE and obtaining the KLiSH clusters, we then compared their performances on benefiting the downstream segmentation task.
We first sampled a synthesized image segmentation dataset for each of the LSE models and KLiSH results, which consisted of 50k images and segmentation masks.
Then, we trained a DeepLabV3 model on each dataset following the training settings described in \cref{sec:segnet_setup}.
Finally, we tested the DeepLabV3 models on the real images from the test split of the CelebAMask-HQ dataset.

\paragraph{Results}\label{subsec:ufgs_results}
We present the qualitative results of our UFGS method in \cref{fig:UFGS}.
As shown in Row 1 and 2, UFGS model segmented facial classes like ``eye'', ``brow'', ``lips'', ``teeth'' and ``nose'' well.
For car images in Row 3 and 4, our UFGS method also successfully segmented ``headlight'', ``body'', ``glasses'', and ``wheel''.
However, we found there was a domain gap between synthesized images and real images.
See Row 1, Col 3 and 4 in \cref{fig:UFGS}.
The hand in front of the face is a rare case in the dataset.
Therefore, the segmentation model cannot deal with this case well.

The quantitative results are reported in \cref{tab:face_iou_global} and \cref{tab:face_iou_classwise}.
As shown in \cref{tab:face_iou_global}, KLiSH surpassed UFGS with AHC by a large margin.
In comparison with few-shot learning methods, UFGS achieved close performance with the one-shot method, LSE, but lagged behind the 10-shot methods. 

In summary, though the performance of our UFGS method was limited by the domain gap as a result of using GANs, we for the first time showed accurate fine-grained semantic segmentation in an unsupervised learning setting.

\subsection{Unsupervised Semantic-Conditional Synthesis}\label{sec:uscs}

Enhancing controllability over the generated images is important for commercial applications of GANs.
Most GANs do not allow the user to control the spatial structure of generated images precisely.
Semantic-Controllable Synthesis (SCS) aims to provide controllability over precise spatial structures of generated images.
Though achieving greater controllability for image synthesis, existing methods for SCS, \eg, pix2pix \cite{Isola2017pix}, SEAN \cite{zhu2020sean} and GauGAN \cite{park2019gaugan}, rely heavily on large-scale human annotated semantic masks.

Recently, the labeling requirements of SCS have been lowered to a few annotations \cite{xu2021linear,ling2021editgan}.
Yet, the user still has to label a few multi-class fine-grained segmentation masks, which limits the application in practice.
For example, users may want to enable SCS on images of tigers.
If there are no semantic segmentation masks on these images, the user needs to annotate 16 to 40 images.
The annotation process would take around 2 to 5 hours, according to Zhang \etal \cite{zhang2021datasetgan}.

To lower the label dependency of SCS methods, we propose the task of Unsupervised SCS, which aims to achieve semantic-controllable image synthesis on arbitrary datasets without human annotations.

\paragraph{Method}
Our USCS application works as follows.
First, the user shall provide a target image dataset.
Then, we train an unconditional GAN and generate an image-segmentation dataset using the same method as described in \cref{subsec:ufgs_mtd}.
Finally, we train a GauGAN \cite{park2019gaugan} on the generated dataset.
The above process can be done offline.
Then users can draw and edit on the semantic mask and the GauGAN will generate desired images. 

We developed a web-based interface to help users generate images with desired structures.
Before the user starts painting, a few images and their semantic masks are shown as examples.
Next, the user can click on a palette and paint with selected semantic classes.
Then, the user can draw freely and submit the painting to the server, which will return an image synthesized given the semantic mask.

\paragraph{Experiment setup}
We conducted the USCS experiment on StyleGAN2 models trained on FFHQ and Car datasets and StyleGAN2-ADA models trained on MetFaces, AFHQ \cite{choi2020stargan} Cat, Dog, and Wildlife split.
Similar to the setting of UFGS, we also synthesized a dataset with 50k paired images and segmentation masks from each GAN and trained a GauGAN on each dataset.
The GauGAN was trained using the official release \footnote{\url{https://www.github.com/nvlabs/spade}} for 10 epochs on 8 GPUs.
The training took around 100 hours for a GauGAN model.

\paragraph{Results}
The editing results on different datasets and GAN models are presented in \cref{fig:uscs}.
We observed that USCS supported precise, diverse, and even out-of-distribution image editing operations.
First, USCS respected the semantic mask well, accurately translating the user's semantic input into corresponding images while maintaining good photo-realisticity.
See the ``big round eye'' example in ADA-Cat.
The enlarged eyes had a matched shape as specified by the user.
Second, our USCS method well demonstrated the diversity and degree of freedom of editing operations.
As shown in the left part of StyleGAN2-FFHQ, the shape of the sunglasses could even be changed to triangle.
Third, the out-of-distribution editing operations could also lead to photorealistic synthesis results.
See the ``elf ear'' examples in StyleGAN2-FFHQ and ADA-Wild, the synthesized sharp ear is not present in the training set of the model.
Besides, the ``blink eye'' in ADA-MetFace and ADA-Wild, and ``more wheels'' in StyleGAN2-Car are all out-of-distribution semantic masks.

To the best of our knowledge, SCS cannot be applied to datasets of wildlife and artistic portraits, \etc, because so far no semantic annotations on such datasets are available.
EditGAN \cite{ling2021editgan} relies on 16 to 30 fine-grained segmentation masks to enable semantic editing on Cat, Dog, \etc, but the cost of labeling is still high (around 10 minutes per image).
In contrast, the USCS model is free from the constraint of annotations.
Please find a demonstration of our USCS web application in {\it Supplementary Video 1} and have a try on our online demonstration\footnote{\url{https://atlantixjj.github.io/KLiSH/}}.

%% file: conclusion.tex
\section{Limitations}\label{sec:limit}

Our method is generally constrained by the performance of GANs.
First, the clustering performance of KLiSH is dependent on the degree of linear separability of GAN's features.
If the GAN does not encode semantics into \emph{one vs.\ rest} linearly separable clusters, then the KLiSH might fail to find meaningful clusters.
Second, the performance of UFGS and USCS is limited by the synthesis quality of the original GAN.
When the GAN fails to generate realistic images, the UFGS model will have a domain gap when applied to real images and USCS will also generate unrealistic images.

\section{Conclusion}\label{sec:conclusion}

In this paper, we propose KLiSH to find semantic clusters in GAN's features.
By making use of GAN's linear separability in semantics, KLiSH achieves better clustering results than conventional clustering algorithms including K-means, AHC, and KASP.
The clusters found by KLiSH accurately correspond to the semantic parts of generated images.
Therefore, for each pretrained GAN, we can generate an image-segmentation dataset for downstream tasks.
One can easily enable semantics-based applications including UFGS and USCS with the dataset generated by GANs.
The surprising performance of UFGS and USCS make us believe that GANs can play a more important role in representation learning.